\documentclass[runningheads]{llncs}
\usepackage[T1]{fontenc}
\usepackage{graphicx}
\usepackage{algorithm,algpseudocode}
\usepackage{times}
\usepackage{amsmath,amssymb,mathtools,stmaryrd,bm}
\usepackage{amsfonts}
\usepackage{url}
\usepackage{booktabs}

\usepackage{wrapfig}
\usepackage{multirow}
\usepackage{array}
\newcolumntype{L}[1]{>{\raggedright\let\newline\\\arraybackslash\hspace{0pt}}m{#1}}
\newcolumntype{C}[1]{>{\centering\let\newline\\\arraybackslash\hspace{0pt}}m{#1}}
\newcolumntype{R}[1]{>{\raggedleft\let\newline\\\arraybackslash\hspace{0pt}}m{#1}}

\newcommand\blfootnote[1]{%
  \begingroup
  \renewcommand\thefootnote{}\footnote{#1}%
  \addtocounter{footnote}{-1}%
  \endgroup
}

\usepackage{xcolor}

\newcommand{\argmax}{\operatornamewithlimits{argmax}} 
\newcommand{\R}{\mathbb{R}} 

\begin{document}
\title{Neural Additive and Basis Models with Feature Selection and Interactions}
%
%
\author{Yasutoshi Kishimoto \and
 Kota Yamanishi \and
 Takuya Matsuda \and
 Shinichi Shirakawa}
\authorrunning{Y. Kishimoto et al.}
%
\institute{Yokohama National University, Kanagawa, Japan}
\maketitle              

\begin{abstract}
Deep neural networks (DNNs) exhibit attractive performance in various fields but often suffer from low interpretability. The neural additive model (NAM) and its variant called the neural basis model (NBM) use neural networks (NNs) as nonlinear shape functions in generalized additive models (GAMs). Both models are highly interpretable and exhibit good performance and flexibility for NN training. NAM and NBM can provide and visualize the contribution of each feature to the prediction owing to GAM-based architectures. However, when using two-input NNs to consider feature interactions or when applying them to high-dimensional datasets, training NAM and NBM becomes intractable due to the increase in the computational resources required. This paper proposes incorporating the feature selection mechanism into NAM and NBM to resolve computational bottlenecks. We introduce the feature selection layer in both models and update the selection weights during training. Our method is simple and can reduce computational costs and model sizes compared to vanilla NAM and NBM. In addition, it enables us to use two-input NNs even in high-dimensional datasets and capture feature interactions. We demonstrate that the proposed models are computationally efficient compared to vanilla NAM and NBM, and they exhibit better or comparable performance with state-of-the-art GAMs.
\blfootnote{Our code is available at \url{https://github.com/shiralab/NAM-FS}.}
\end{abstract}

\section{Introduction}
Deep neural networks (DNNs) have become a standard tool in artificial intelligence owing to their high representation ability and flexible model training. DNNs have shown remarkable performance, such as in computer vision and natural language processing, and have become a promising model even for tabular datasets~\cite{tabnet,Gorishniy2021,NODE}. A well-known drawback of DNNs is the low interpretability and explainability of the prediction mechanism due to multi-layered nonlinear computations, making them difficult to use in applications that require high reliability, such as medical and legal applications.
Numerous methods have been examined to interpret trained DNNs, including locally approximating the decision boundary by an interpretable model~\cite{lime} and visualizing important input image regions using gradient information~\cite{grad-cam}. However, these methods do not explain the global behavior of the models, and their explanations are often not faithful~\cite{rudin2019stop}. The alternative choice is to adopt interpretable models, such as shallow decision trees and linear models, which often suffer from low prediction accuracy.

The neural additive model (NAM)~\cite{nam} is a type of generalized additive model (GAM) that possesses high interpretability and good prediction accuracy. GAMs are formed as the sum of nonlinear transformations for each input feature, referred to as shape functions. NAM uses one-input and one-output neural networks (NNs) as shape functions in GAMs and jointly trains all NNs through backpropagation. Because interactions between features do not exist in the shape functions, the global behavior of NAMs can be easily interpreted by visualizing shape functions. 
The neural basis model (NBM)~\cite{NBM} is another type of GAM, which uses a single shared NN for the basis of shape functions. The number of parameters in NBM can be significantly smaller than that in NAM because of the shared NN. Additionally, the implementation of NBM is simpler than that of NAM, making it easy to realize efficient implementation.
Thanks to the NN shape functions, NAM and NBM can represent complex shape functions, which contributes to good prediction accuracy. They can also exploit the flexibility of DNN training, for example, extending it to multi-task or multi-label learning~\cite{nam}.

To enhance the prediction performance of GAMs without losing interpretability, adding pairwise interactions (two input shape functions) is promising. Such models are called GAMs plus interactions (GA$^2$Ms)~\cite{ga2m}. Because the global behavior of two-input shape functions, such as a two-input NN, can be visualized using heat maps, GA$^2$Ms maintain high interpretability and are expected to improve performance. The naive extension method of NAM and NBM to consider pairwise feature interactions involves preparing the shape functions for all possible feature pairs. However, such a naive extension for high-dimensional datasets is intractable in NAM because the number of NNs and trainable parameters increases, leading to failure to run. Although the increase in the number of parameters of NBM is not as significant as that of NAM, the increase in the calculation cost becomes intractable. Even if only one-input NNs are used, the computations of NAM and NBM are still slow for high-dimensional ($>1,000$) datasets. Additionally, increasing the shape functions may also compromise the interpretability of NAMs because considering and visualizing a large number of shape functions becomes unmanageable.

We incorporate a simple feature selection mechanism into NAM and NBM to reduce the number of shape functions and resolve their computational bottlenecks. The proposed method determines the features for one- and two-input NNs during model training.
The feature selection allows us to reduce the computational complexity of NAM and NBM, making it possible to train NAM and NBM on high-dimensional datasets and easily incorporate pairwise interactions. We also discuss the model complexities of our models. We then compare the prediction performance of our models, termed NAM-FS and NBM-FS, with that of existing GAMs, other interpretable models, and non-interpretable but accurate models using high-dimensional classification datasets. The experimental results demonstrate that our models perform better or are comparable to other existing GAMs and GA$^2$Ms. We further compare the prediction performance of our models to vanilla NAM and NBM with pre-selected features by mutual information-based feature selection. Our models perform better than NAM and NBM with pre-selected features, demonstrating the effectiveness of the feature selection during model training.

Our contributions are as follows. (i) We incorporate a simple feature selection method into NAM and NBM, which enables us to apply them to high-dimensional datasets and consider pairwise interactions. (ii) We show that our proposed models achieve better throughput than vanilla NAM and NBM on high dimensions and better or comparable performance to other GAMs.

\section{Generalized Additive Models (GAMs)} \label{sec:GAM}
This section describes GAMs~\cite{gam2012}, NAM~\cite{nam}, and NBM~\cite{NBM}. GAMs are highly transparent due to their structure, which prepares one nonlinear function per input feature and uses the sum of the outputs of the nonlinear functions as the model's prediction. Given a $D$-dimensional input vector $\bm{x} \in \R^D$, the form of GAMs with a single output is given by $y (\bm{x}) = \sum_{i=1}^{D} f_i (x_i) + b$, where $f_i$ is called the shape function of the $i$-th feature, which is an univariate nonlinear function, $x_i$ is the $i$-th element of $\bm{x}$, and $b$ is the bias term.
Because each shape function value is decided by a single feature and the shape function values are added to make the model's output, the shape function value can be considered as its contribution to the model's output. Owing to this structure, we can interpret how each feature contributes to the model's prediction by showing the output of each nonlinear model. However, GAM cannot make predictions considering the interactions between features, which may lead to a degradation in prediction accuracy compared to complicated models.

Generalized additive models plus interactions (GA$^2$Ms)~\cite{ga2m} are advanced versions of GAMs that consider interactions between features by adding bivariate shape functions to GAMs. The form of GA$^2$M is expressed as
\begin{equation}
  \label{eq:ga2m}
  y (\bm{x}) = \sum_{i=1}^{D} f_i (x_i) + \sum_{i=1}^{D} \sum_{j > i}^{D} f_{ij} (x_i, x_j) + b \enspace ,
\end{equation}
where $f_{ij}$ is the bivariate nonlinear shape function of the ($i$, $j$)-feature pair. For two-input shape functions, we can visualize the input-output relationship using a heat map to interpret the prediction mechanism. For all interactions, there are $D(D-1)/2$ possible pairs in the $D$ features; thus, the computational complexity increases with the number of features.

When applying GAMs to tasks with multiple outputs, such as multiclass classification, we can linearly combine shape functions to obtain multiple outputs as in \cite{NBM}. The prediction function $y_l$ for the $l$-th output is given by
\begin{equation}
  \label{eq:multiclass-gam}
  y _l(\bm{x}) = \sum_{i=1}^{D} f_i (x_i) w_{li} + b_l \enspace ,
\end{equation}
where $w_{li}$ and $b_l$ indicate the coefficient of $i$-the shape function and the bias term for the $l$-th output, respectively. For GA$^2$M, we also introduce the coefficient for each bivariate shape function for the ($i$, $j$)-feature pair to support multiple outputs.

Several versions of GAMs and GA$^2$Ms have been proposed. Explainable boosting machine (EBM)~\cite{ebm-implement} is a tree-based GAM and GA$^2$M. NODE-GAM~\cite{NODE-GAM} is based on neural oblivious decision ensembles (NODE)~\cite{NODE} that are ensembles of oblivious decision trees with gradient-based training. In the following, we describe the neural network versions of GAMs as our baseline models.

\subsection{Neural Additive Model (NAM)} \label{sec:NAM}
NAM~\cite{nam}, a variant of GAMs, uses NNs as nonlinear shape functions $f_i$. For the GAM form, a one-input and one-output NN $f_i$ with trainable parameters is used for the $i$-th shape function. The training of NAM is performed by stochastic gradient descent (SGD), as in standard NNs, and all NNs composing the NAM are trained simultaneously. In NAM, the feature dropout and the output penalty are used for regularization, in addition to the standard dropout~\cite{dropout} and the weight decay for individual NN. In feature dropout, the dropout technique is applied to the NN outputs, i.e., the shape function outputs. The output penalty is the mean of the squares of the NN outputs. Due to the end-to-end learning manner, NAMs inherit the flexibility of NNs.
NAM can be extended to consider pairwise interactions by adding two-input NNs for all possible feature pairs as bivariate shape functions $f_{ij}$. We denote NAM with pairwise interactions as NA$^2$M. However, the number of parameters and the computational cost increase by $O(D^2)$ as the number of features increases in NA$^2$M.

\subsection{Neural Basis Model (NBM)} \label{sec:NBM}
NBM~\cite{NBM} uses multiple basis functions shared across all features to make predictions. NBM decomposes each shape function $f_i$ into a linear combination of a small set of basis functions that are shared among all shape functions. NBM uses only a single one-input and $B$-output NN to represent the shared basis functions. The training of NBM is performed by SGD. Due to the shared basis functions, the number of parameters in NBM can be reduced compared to NAM. The shape function in NBM is given by
\begin{equation}
    \label{eq:nbm-shape}
    f_i (x_i) = \sum_{k=1}^{B} h_k (x_i) a_{ik} \enspace ,
\end{equation}
where $\{h_1, h_2, \dots, h_B \}$ denotes a set of $B$ basis functions represented by a one-input and $B$-output NN, and $a_{ik}$ are the coefficients of $k$-th basis function. When extending NBM to the form of GA$^2$M, the two-input shape function $f_{ij}$ in \eqref{eq:ga2m} is given by
\begin{equation}
    \label{eq:nb2m-shape}
    f_{ij} (x_i, x_j) = \sum_{k=1}^{B} u_k (x_i, x_j) c_{ijk} \enspace ,
\end{equation}
where $\{u_1, u_2, \dots, u_B \}$ denotes a set of $B$ bivariate basis functions represented by a two-input and $B$-output NN, and $c_{ijk}$ are the coefficients of $k$-th output of NN. We denote NBM with pairwise interactions as NB$^2$M.

\section{NAM and NBM with Feature Selection} \label{sec:proposed}
\subsection{Motivation}
Although NAM and NBM possess promising interpretability, prediction performance, and flexible training schemes, there is a problem when applying them to high-dimensional datasets.\footnote{Although high-dimensional datasets with sparse features can be handled by NBM with the specialized implementation, as shown in \cite{NBM}, it cannot be applied to dense features. In our experiments, NA$^2$M and NB$^2$M could not run on more than hundred features, and training NAM and NBM slowed down on more than thousand features in dense feature datasets.} In NAM, preparing NNs for all features brings a disadvantageous increase in the number of parameters and computational cost, which is notable in NA$^2$M. Although the increase in the number of parameters in NBM can be relaxed due to the shared basis functions, NBM still results in an increase in the calculation cost in (dense) high-dimensional datasets. In addition, the large number of shape functions to be managed are intractable for visualizing and interpreting the models' behavior. Reducing the number of single features and pairs of features in NAM and NBM will be useful in applying them to high-dimensional datasets.
Therefore, we incorporate feature selection layers using entmax~\cite{entmax} into NAM and NBM to select the predefined numbers of single features and pairs of features. The feature selection is performed through model training by optimizing the feature selection weights. That is, our method inherits the end-to-end learning properties of NAM and NBM. Our method is simple, but it is valuable to develop scalable NAM and NBM for high-dimensional datasets. It should be noted that our method does not compromise the interpretability of NAM and NBM.

\subsection{Model Architecture}
We prepare $K_1$ one-input shape functions $\{f_1, \dots, f_{K_1}\}$ and $K_2$ two-input shape functions $\{g_1, \dots, g_{K_2}\}$, where the total number of shape functions is $K = K_1 + K_2$. Let us consider an additive model with interaction terms, $\tilde{y}: \R^{K_1} \times \R^{K_2} \times \R^{K_2} \to \R$, which takes $K_1 + 2 K_2$ input variables as
\begin{equation}
    \label{eq:gam-fs}
    \tilde{y} (\bm{z}_1, \bm{z}_2, \bm{z}_3) = \sum_{i=1}^{K_1} f_i \left(z_{1,i} \right) + \sum_{i=1}^{K_2} g_i \left(z_{2, i}, \; z_{3, i} \right) + b \enspace ,
\end{equation}
where $\bm{z}_1 \in \R^{K_1}$ and $\bm{z}_2, \bm{z}_3 \in \R^{K_2}$ are the input variables for one- and two-input shape functions, respectively, $z_{j,i}$ indicates the $i$-th element of $\bm{z}_j$, and $b$ is a bias term.
The form of \eqref{eq:gam-fs} can reduce the number of shape functions to be managed compared to \eqref{eq:ga2m} by setting $K_1 < D$ for univariate functions and $K_2 < D(D-1)/2$ for bivariate functions. Setting $K_2=0$ means that the model only considers univariate effects, whereas $K_2>0$ can consider interactions of $K_2$ feature pairs.\footnote{Our method can be extended to three or more input shape functions to capture high-order feature interactions while it compromises the interpretability.}

Let us consider selecting $K_1 + 2K_2$ features from $D$ to be fed into \eqref{eq:gam-fs}. To maintain end-to-end learning, we consider differentiable feature selection that enables us to select features through SGD-based training. Given a one-hot vector $\bar{\bm{v}} \in \{0, 1 \}^D$, the selection of a feature from a $D$-dimensional feature vector $\bm{x}$ can be represented by the dot product of $\bar{\bm{v}} \cdot \bm{x}$. As $\bar{\bm{v}}$ is one-hot and discrete, optimizing it directly using a gradient method is not feasible. Therefore, we replace it with continuous variables given by entmax~\cite{entmax} to make feature selection differentiable. Entmax is the generalized version of the softmax and sparsemax functions, and it has a hyperparameter $\alpha$ used to tune the sparsity of the output probability. Entmax is equivalent to softmax and sparsemax if $\alpha = 1$ and $\alpha = 2$, respectively. We set $\alpha = 1.5$ in our experiments. We also introduce the temperature parameter $\tau$ and anneal it in model training to make the entmax values approach one-hot, similar to \cite{NODE-GAM}. Note that other continuous relaxation approaches can be used in our method, such as the Gumbel-Softmax~\cite{gumbel-softmax,concrete-distribution} function.

Denoting the value of entmax with trainable logit parameters $\bm{\pi} \in \R^{D}$ as $\bm{v} = \mathrm{entmax}_{\alpha} (\bm{\pi} / \tau) \in \Delta^{D-1}$, where $\Delta^{D-1}$ is the standard simplex in $\R^{D}$, we get the relaxed feature selection process as $\bm{v} \cdot \bm{x}$. With a low-temperature parameter $\tau$, the vector $\bm{v}$ is expected to approach one-hot, and we can pick up one feature from $\bm{x}$. By using this feature selection relaxation, we can get our additive model with feature selection as follows:
\begin{equation}
    \label{eq:gam-fs2}
    y (\bm{x}) = \sum_{i=1}^{K_1} f_i \left( \bm{v}_{1,i} \cdot \bm{x} \right) + \sum_{i=1}^{K_2} g_j \left( \bm{v}_{2,i} \cdot \bm{x}, \; \bm{v}_{3,i} \cdot \bm{x} \right) + b \enspace ,
\end{equation}
where $\{ \bm{v}_{1,i} \}_{i=1}^{K_1}$, $\{ \bm{v}_{2,i} \}_{i=1}^{K_2}$, and $\{ \bm{v}_{3,i} \}_{i=1}^{K_2}$ indicate the entmax values from different trainable logit parameters.

The form of \eqref{eq:gam-fs2} can be extended to multiple outputs, e.g. multiclass classification, as in \eqref{eq:multiclass-gam}. By introducing the coefficients for each shape function, $\{w^{(1)}_{li}\}_{i=1}^{K_1}$ and $\{w^{(2)}_{li}\}_{i=1}^{K_2}$, we obtain the $l$-th output of the model as
\begin{equation}
  \label{eq:multiclass-gam-fs}
  y_l (\bm{x}) = \sum_{i=1}^{K_1} f_i \left( \bm{v}_{1,i} \cdot \bm{x} \right) w^{(1)}_{li} + \sum_{i=1}^{K_2} g_j \left( \bm{v}_{2,i} \cdot \bm{x}, \; \bm{v}_{3,i} \cdot \bm{x} \right) w^{(2)}_{li} + b_l \enspace .
\end{equation}

In model training, all trainable parameters in the NN shape functions and entmax functions are optimized using SGD as the usual NN training. We gradually decrease the temperature parameter $\tau$ during model training. When the temperature parameter $\tau$ becomes sufficiently small, the input features of shape functions are fixed based on the logit parameters $\bm{\pi}$ in the entmax function, meaning that the calculation of $\bm{v} \cdot \bm{x}$ can be ignored. We select the feature that corresponds to the largest logit parameter in these phases, that is, an input feature $z_i$ becomes $x_k$, where $k = \argmax_j \pi_{i, j}$. In the inference phase, we also select the features in the same way.

\subsubsection{NAM with Feature Selection (NAM-FS)}
We call the additive model of \eqref{eq:gam-fs2} or \eqref{eq:multiclass-gam-fs} using NNs as shape functions NAM-FS. The model is denoted as NAM-FS$_{(K_1)}$ when $K_2=0$, that is, no feature interactions, and as NA$^2$M-FS$_{(K_1, K_2)}$ when considering feature interactions ($K_2>0$).

\subsubsection{NBM with Feature Selection (NBM-FS)}
Similar to NAM-FS, we call the additive model of \eqref{eq:gam-fs2} or \eqref{eq:multiclass-gam-fs} using the shape functions given by \eqref{eq:nbm-shape} and \eqref{eq:nb2m-shape} NBM-FS. The model is denoted as NBM-FS$_{(K_1)}$ when $K_2=0$ and NB$^2$M-FS$_{(K_1, K_2)}$ when $K_2>0$.

\subsection{Implementation Remark}
The actual computational cost significantly depends on the implementation and library used. Radenovic et al.~\cite{NBM} reimplemented NAM and optimized the computation using grouped 1-D convolution in PyTorch~\cite{pytorch} We also used the PyTorch library (version 1.13.1) and further improved the implementation of NAM. We found that using the \texttt{torch.bmm} function (batch matrix-matrix product) to implement MLP layers in NAM is more efficient than using the 1-D convolution function. Our implementation of the NAM layer speeds up by approximately $\times 2$--$\times 6$ compared to when using the 1-D convolution function on an NVIDIA A100 GPU machine.

\section{Discussion of Model Complexities} \label{sec:complexity}
While NAM requires as many NNs as the number of features $D$, NAM-FS requires only $K_1$ NNs and $K_1 D$ parameters for feature selection. The number of parameters in NAM-FS is significantly small when setting $K_1 \ll D$. In the case of NA$^2$M, $D + D(D-1)/2$ NNs are required, which results in a significant increase in the number of parameters and computational cost as $D$ increases. However, the increase in the number of parameters in our NA$^2$M-FS can be reduced because the number of NNs in NA$^2$M-FS is $K_1 + K_2$. In the NAM family, the total multiply-accumulate operations (MACs) in the forward calculation, which affect the computational cost, are equivalent to the number of parameters.
In NBMs, the number of parameters does not increase significantly compared to NAMs as $D$ increases, while MACs increase. Due to the increase in MACs, NBM cannot run in the case of a large feature dimension and when considering pairwise interactions. Our NBM-FS and NB$^2$M-FS can save MACs because the calculations are required only for $K_1$ and $K_1 + K_2$ shape functions, respectively.

\begin{figure}[bt]
\begin{minipage}[]{0.48\columnwidth}
    \centering
    \includegraphics[keepaspectratio, width=\columnwidth]{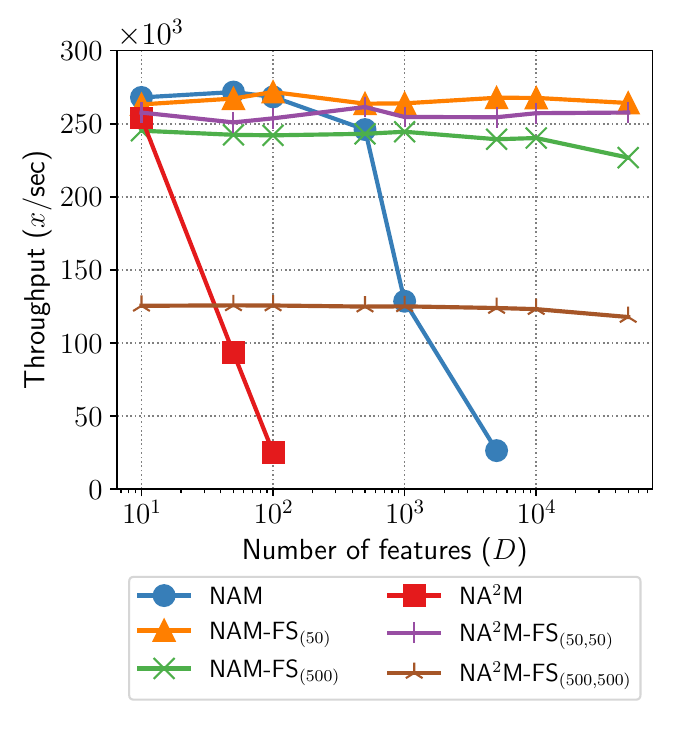}
    \caption{Throughput (the number of inputs processed per second) of NAM, NAM-FS, NA$^2$M, and NA$^2$M-FS}
    \label{fig:nam-thrpt}
  \end{minipage}
  \hspace{0.03\columnwidth}
  \begin{minipage}[]{0.48\columnwidth}
    \centering
    \includegraphics[keepaspectratio, width=\columnwidth]{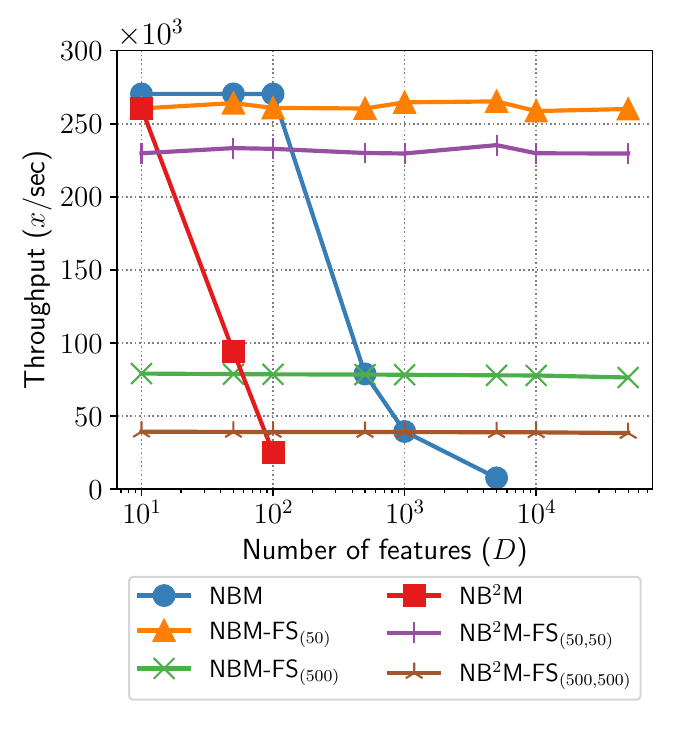}
    \caption{Throughput  (the number of inputs processed per second) of NBM, NBM-FS, NB$^2$M, and NB$^2$M-FS}
    \label{fig:nbm-thrpt}
  \end{minipage}
\end{figure}

We then measured the throughput rate, defined as the number of data instances processed per second (denoting $\bm{x}$/sec), in order to check the actual computational cost. The throughput directly affects the training time. The NN architectures in NAM and NBM are the same as those used in Section \ref{sec:experiment}. The batch size was set to $1,024$, and the numbers of one- and two-input shape functions were set to $K_1 \in \{ 50, 500\}$ for NAM-FS and NBM-FS and to $(K_1, K_2) \in \{(50, 50), (500, 500) \}$ for NA$^2$M-FS and NB$^2$M-FS. We ran each model on an NVIDIA A100 GPU (80GB memory) machine. Figures \ref{fig:nam-thrpt} and \ref{fig:nbm-thrpt} show the throughput ($\bm{x}$/sec) of the NAM and NBM families for forward calculation in training mode, respectively. We observe that the throughputs of NA$^2$M and NB$^2$M rapidly worsen as $D$ increases, and they cannot run on more than 100 features due to the huge memory consumption. Even in NAM and NBM, the throughput slows down for more than 1,000 features.
In contrast, our models are not affected by the increase in the number of features. Figure \ref{fig:nam-thrpt} shows that NAM-FS$_{(50)}$, NAM-FS$_{(500)}$, and NA$^2$M-FS$_{(50, 50)}$ exhibit good throughput for any $D$ (up to $5 \times 10^4$). NA$^2$M-FS$_{(500, 500)}$ was slower than NA$^2$M-FS$_{(50, 50)}$ due to the large settings of $K_1$ and $K_2$. However, it can run on more than a hundred features while considering feature interactions of 500 pairs. Figure \ref{fig:nbm-thrpt} shows a similar trend to that in Figure \ref{fig:nam-thrpt}. In summary, NAM-FS and NBM-FS are advantageous in terms of throughput with more than 1,000 features, and NA$^2$M-FS and NB$^2$M-FS enable us to exploit feature interactions in datasets with more than 100 features that vanilla NA$^2$M and NB$^2$M cannot handle.

\section{Experiments} \label{sec:experiment}
We compared the performance of our methods with that of other GAMs and well-known models on high-dimensional datasets ($D=500$ to $5000$) that cannot run NA$^2$M and NB$^2$M due to high computational costs. We further evaluated our feature-selection mechanism by comparing its performance with those of vanilla NAM and NBM using pre-selected features. We used the six classification datasets listed in Table \ref{tab:dataset}. For F-MNIST and Epsilon, we used the test data provided in the datasets. We randomly split the data into 80\% and 20\% for training and testing, respectively, for the remaining datasets. Each data was standardized using the training dataset.

\subsection{Experimental Settings}
The NN architectures used in the NAM and NBM families were determined by referring to original papers~\cite{nam,NBM}. We used multi-layer perceptrons (MLPs) containing three hidden layers with (64, 64, 32) units and those with (256, 128, 128) for the NAM and NBM families, respectively. Batch normalization and ReLU activation were applied to each layer. The number of bases was set to $B=100$ for NBM and NBM-FS, and $B=200$ for NB$^2$M-FS by following the default setting in \cite{NBM}. We varied the number of shape functions, $K_1$ and $K_2$, in our models as $(K_1, K_2) \in \{ (50, 0), (500, 0), (50, 50), (500, 500) \}$ to investigate their impact on performance. Our models with $K_2 > 0$ can capture pairwise feature interactions.

We used cross-entropy loss and Adam optimizer~\cite{Kingma2015} with an initial learning rate of $0.01$. For the training procedure, we early stopped the training if the validation accuracy did not improve for 11,000 iterations, and we decayed the learning rate to 1/5 if the validation accuracy did not improve for 5,000 iterations. This training procedure was based on \cite{NODE-GAM} and was applied to all variants of NAM and NBM. The mini-batch size was set to 1,024. For the annealing of the temperature parameter in the proposed method, we linearly decreased it from $1$ to $0.01$ during the first 4,000 iterations to make the feature selection weights one-hot. After 4,000 iterations, we fixed the features to be fed into NNs and continued to train the model. 

Other hyperparameters, dropout rate, weight decay coefficient, output penalty, feature dropout in the NAM family, basis dropout in the NBM family, were tuned by grid search. We divided the training dataset into 90\% and 10\% for the model training and validation data, respectively. The validation data were used for hyperparameter tuning and early stopping. By using tuned hyperparameters, we report the average test accuracy and standard deviation over 10 runs with different random seeds. Each NAM- and NBM-based model was run on a single NVIDIA A100 GPU (40GB or 80GB of memory).

\begin{table}[tb]
  \caption{Summary of datasets. \#Feat, \#Train, \#Test, and \#Class indicate the numbers of features, training data, test data, and classes.}
  \label{tab:dataset}
  \centering
  \begin{tabular}{lrrrr}
    \toprule
    \textbf{Dataset} & \textbf{\#Feat} & \textbf{\#Train} & \textbf{\#Test} & \textbf{\#Class} \\
    \midrule
     HAR\cite{HAR}     & 561 & 8,239 & 2,060 & 6 \\
     ISOLET~\cite{ISOLET}  & 617 & 6,237 & 1,560 & 26 \\
     F-MNIST~\cite{FMNIST} & 784 & 60,000 & 10,000 & 10 \\
     Epsilon~\cite{Epsilon} & 2,000 & 400,000 & 100,000 & 2 \\
     guillermo~\cite{Guillermo}  & 4,296 & 16,000 & 4,000 & 2 \\
     Gisette~\cite{gisette}   & 5,000 & 5,600 & 1,400 & 2 \\
     \bottomrule
  \end{tabular}
\end{table}

\begin{table*}[tb]
  \def\cw{1.45cm}
  \def\hs{1mm}
  \centering
  \caption{Comparison of test accuracy with baselines. Models in the first and second groups belong to GAMs and GA$^2$M, respectively. The \textbf{bold} font indicates the model with the highest accuracy among GAMs or GA$^2$Ms. The \underline{underline} means the best-performing model in considered models.}
  \label{tab:comp_baseline}
  \begin{tabular}{L{2.5cm}C{\cw}C{\cw}C{\cw}C{\cw}C{\cw}C{\cw}}
    \toprule
    \textbf{Model} & \textbf{HAR} & \textbf{ISOLET} & \textbf{F-MNIST} & \textbf{Epsilon} & \textbf{guillermo} & \textbf{Gisette} \\
    \midrule
    NAM &
		0.9850 \tiny{\textcolor{gray}{$\pm$0.0021}} &
		0.9534 \tiny{\textcolor{gray}{$\pm$0.0039}} &
		0.8634 \tiny{\textcolor{gray}{$\pm$0.0018}} &
		\textbf{0.8950} \tiny{\textcolor{gray}{$\pm$0.0008}} &
		0.7714 \tiny{\textcolor{gray}{$\pm$0.0071}} &
		0.9715 \tiny{\textcolor{gray}{$\pm$0.0053}} \\
	NAM-FS$_{(50)}$ &
		0.9764 \tiny{\textcolor{gray}{$\pm$0.0035}} &
		0.9257 \tiny{\textcolor{gray}{$\pm$0.0108}} &
		0.8203 \tiny{\textcolor{gray}{$\pm$0.0047}} &
		0.8443 \tiny{\textcolor{gray}{$\pm$0.0054}} &
		0.7545 \tiny{\textcolor{gray}{$\pm$0.0052}} &
		0.9438 \tiny{\textcolor{gray}{$\pm$0.0064}} \\
	NAM-FS$_{(500)}$ &
		0.9877 \tiny{\textcolor{gray}{$\pm$0.0024}} &
		\textbf{0.9580} \tiny{\textcolor{gray}{$\pm$0.0027}} &
		0.8601 \tiny{\textcolor{gray}{$\pm$0.0018}} &
		0.8927 \tiny{\textcolor{gray}{$\pm$0.0007}} &
		\textbf{0.7802} \tiny{\textcolor{gray}{$\pm$0.0050}} &
		0.9684 \tiny{\textcolor{gray}{$\pm$0.0048}} \\
	NBM &
		\textbf{0.9879} \tiny{\textcolor{gray}{$\pm$0.0024}} &
		0.9494 \tiny{\textcolor{gray}{$\pm$0.0029}} &
		\textbf{0.8668} \tiny{\textcolor{gray}{$\pm$0.0014}} &
		0.8943 \tiny{\textcolor{gray}{$\pm$0.0027}} &
		0.7280 \tiny{\textcolor{gray}{$\pm$0.0183}} &
		0.9711 \tiny{\textcolor{gray}{$\pm$0.0033}} \\
	NBM-FS$_{(50)}$ &
		0.9739 \tiny{\textcolor{gray}{$\pm$0.0051}} &
		0.9297 \tiny{\textcolor{gray}{$\pm$0.0033}} &
		0.8212 \tiny{\textcolor{gray}{$\pm$0.0027}} &
		0.8499 \tiny{\textcolor{gray}{$\pm$0.0038}} &
		0.7564 \tiny{\textcolor{gray}{$\pm$0.0104}} &
		0.9509 \tiny{\textcolor{gray}{$\pm$0.0076}} \\
	NBM-FS$_{(500)}$ &
		0.9858 \tiny{\textcolor{gray}{$\pm$0.0025}} &
		0.9547 \tiny{\textcolor{gray}{$\pm$0.0051}} &
		0.8618 \tiny{\textcolor{gray}{$\pm$0.0021}} &
		0.8948 \tiny{\textcolor{gray}{$\pm$0.0005}} &
		0.7739 \tiny{\textcolor{gray}{$\pm$0.0114}} &
		\textbf{0.9740} \tiny{\textcolor{gray}{$\pm$0.0024}} \\
	EBM &
		0.9835 \tiny{\textcolor{gray}{$\pm$0.0007}} &
		0.9493 \tiny{\textcolor{gray}{$\pm$0.0009}} &
		0.8639 \tiny{\textcolor{gray}{$\pm$0.0007}} &
		0.8788 \tiny{\textcolor{gray}{$\pm$0.0002}} &
		0.7440 \tiny{\textcolor{gray}{$\pm$0.0039}} &
		0.9575 \tiny{\textcolor{gray}{$\pm$0.0013}} \\
	NODE-GAM &
		0.9845 \tiny{\textcolor{gray}{$\pm$0.0028}} &
		0.9438 \tiny{\textcolor{gray}{$\pm$0.0047}} &
		0.8621 \tiny{\textcolor{gray}{$\pm$0.0022}} &
		0.8945 \tiny{\textcolor{gray}{$\pm$0.0004}} &
		0.7710 \tiny{\textcolor{gray}{$\pm$0.0068}} &
		0.9706 \tiny{\textcolor{gray}{$\pm$0.0031}} \\
		\midrule
	NA$^2$M-FS$_{(50, 50)}$ &
		0.9859 \tiny{\textcolor{gray}{$\pm$0.0024}} &
		0.9471 \tiny{\textcolor{gray}{$\pm$0.0036}} &
		0.8634 \tiny{\textcolor{gray}{$\pm$0.0022}} &
		0.8772 \tiny{\textcolor{gray}{$\pm$0.0015}} &
		0.7908 \tiny{\textcolor{gray}{$\pm$0.0094}} &
		0.9665 \tiny{\textcolor{gray}{$\pm$0.0027}} \\
	NA$^2$M-FS$_{(500, 500)}$ &
		\textbf{0.9900} \tiny{\textcolor{gray}{$\pm$0.0016}} &
		\textbf{0.9645} \tiny{\textcolor{gray}{$\pm$0.0031}} &
		\textbf{0.8949} \tiny{\textcolor{gray}{$\pm$0.0017}} &
		0.8931 \tiny{\textcolor{gray}{$\pm$0.0003}} &
		\textbf{0.8012} \tiny{\textcolor{gray}{$\pm$0.0026}} &
		0.9611 \tiny{\textcolor{gray}{$\pm$0.0356}} \\
	NB$^2$M-FS$_{(50, 50)}$ &
		0.9816 \tiny{\textcolor{gray}{$\pm$0.0034}} &
		0.9482 \tiny{\textcolor{gray}{$\pm$0.0064}} &
		0.8606 \tiny{\textcolor{gray}{$\pm$0.0038}} &
		0.8826 \tiny{\textcolor{gray}{$\pm$0.0019}} &
		0.7892 \tiny{\textcolor{gray}{$\pm$0.0074}} &
		0.9597 \tiny{\textcolor{gray}{$\pm$0.0048}} \\
	NB$^2$M-FS$_{(500, 500)}$ &
		0.9863 \tiny{\textcolor{gray}{$\pm$0.0022}} &
		0.9597 \tiny{\textcolor{gray}{$\pm$0.0033}} &
		0.8925 \tiny{\textcolor{gray}{$\pm$0.0025}} &
		\textbf{0.8946} \tiny{\textcolor{gray}{$\pm$0.0007}} &
		0.7702 \tiny{\textcolor{gray}{$\pm$0.0059}} &
		\textbf{0.9759} \tiny{\textcolor{gray}{$\pm$0.0019}} \\
	EB$^2$M &
		N/A &
		N/A &
		N/A &
		0.8794 \tiny{\textcolor{gray}{$\pm$0.0003}} &
		0.7834 \tiny{\textcolor{gray}{$\pm$0.0024}} &
		0.9671 \tiny{\textcolor{gray}{$\pm$0.0011}} \\
	NODE-GA$^2$M &
		0.9867 \tiny{\textcolor{gray}{$\pm$0.0031}} &
		0.8876 \tiny{\textcolor{gray}{$\pm$0.0541}} &
		0.8817 \tiny{\textcolor{gray}{$\pm$0.0016}} &
		0.8941 \tiny{\textcolor{gray}{$\pm$0.0003}} &
		0.7861 \tiny{\textcolor{gray}{$\pm$0.0050}} &
		0.9721 \tiny{\textcolor{gray}{$\pm$0.0039}} \\
		\midrule
	LR &
		0.9820 \tiny{\textcolor{gray}{$\pm$0.0000}} &
		0.9583 \tiny{\textcolor{gray}{$\pm$0.0000}} &
		0.8470 \tiny{\textcolor{gray}{$\pm$0.0000}} &
		\underline{0.8990} \tiny{\textcolor{gray}{$\pm$0.0000}} &
		0.7203 \tiny{\textcolor{gray}{$\pm$0.0000}} &
		\underline{0.9786} \tiny{\textcolor{gray}{$\pm$0.0000}} \\
	Decision Tree &
		0.9388 \tiny{\textcolor{gray}{$\pm$0.0023}} &
		0.8158 \tiny{\textcolor{gray}{$\pm$0.0027}} &
		0.8067 \tiny{\textcolor{gray}{$\pm$0.0008}} &
		0.6826 \tiny{\textcolor{gray}{$\pm$0.0000}} &
		0.7588 \tiny{\textcolor{gray}{$\pm$0.0006}} &
		0.9433 \tiny{\textcolor{gray}{$\pm$0.0027}} \\
	MLP &
		0.9852 \tiny{\textcolor{gray}{$\pm$0.0024}} &
		\underline{0.9663} \tiny{\textcolor{gray}{$\pm$0.0026}} &
		\underline{0.9099} \tiny{\textcolor{gray}{$\pm$0.0024}} &
		0.8970 \tiny{\textcolor{gray}{$\pm$0.0005}} &
		0.7738 \tiny{\textcolor{gray}{$\pm$0.0043}} &
		0.9732 \tiny{\textcolor{gray}{$\pm$0.0025}} \\
	XGBoost &
		\underline{0.9907} \tiny{\textcolor{gray}{$\pm$0.0013}} &
		0.9567 \tiny{\textcolor{gray}{$\pm$0.0029}} &
		0.8979 \tiny{\textcolor{gray}{$\pm$0.0023}} &
		0.8836 \tiny{\textcolor{gray}{$\pm$0.0006}} &
		\underline{0.8195} \tiny{\textcolor{gray}{$\pm$0.0040}} &
		0.9711 \tiny{\textcolor{gray}{$\pm$0.0021}} \\
    \bottomrule
  \end{tabular}
\end{table*}

\begin{table*}[tb]
  \def\cw{1.5cm}
  \def\hs{1mm}
  \centering
  \caption{Test accuracy of NAM and NBM with pre-selected features. The \underline{underline} means that its model is better than the corresponding NAM-FS$_{(K_1)}$ or NBM-FS$_{(K_1)}$.}
  \label{tab:prefs}
  \begin{tabular}{lC{\cw}C{\cw}C{\cw}C{\cw}C{\cw}C{\cw}}
    \toprule
    \textbf{Model} & \textbf{HAR} & \textbf{ISOLET} & \textbf{F-MNIST} & \textbf{Epsilon} & \textbf{guillermo} & \textbf{Gisette} \\
    \midrule
    NAM$_{(50)}$ &
		0.9281 \tiny{\textcolor{gray}{$\pm$0.0055}} &
		0.7474 \tiny{\textcolor{gray}{$\pm$0.0065}} &
		0.5665 \tiny{\textcolor{gray}{$\pm$0.0117}} &
		0.8007 \tiny{\textcolor{gray}{$\pm$0.0005}} &
		\underline{0.7678} \tiny{\textcolor{gray}{$\pm$0.0031}} &
		0.9053 \tiny{\textcolor{gray}{$\pm$0.0037}} \\
	NAM$_{(500)}$ &
		0.9814 \tiny{\textcolor{gray}{$\pm$0.0020}} &
		\underline{0.9587} \tiny{\textcolor{gray}{$\pm$0.0034}} &
		\underline{0.8613} \tiny{\textcolor{gray}{$\pm$0.0014}} &
		0.8839 \tiny{\textcolor{gray}{$\pm$0.0005}} &
		0.7256 \tiny{\textcolor{gray}{$\pm$0.0587}} &
		\underline{0.9709} \tiny{\textcolor{gray}{$\pm$0.0028}} \\
	NBM$_{(50)}$ &
		0.9246 \tiny{\textcolor{gray}{$\pm$0.0022}} &
		0.7535 \tiny{\textcolor{gray}{$\pm$0.0068}} &
		0.5901 \tiny{\textcolor{gray}{$\pm$0.0023}} &
		0.7995 \tiny{\textcolor{gray}{$\pm$0.0004}} &
		\underline{0.7927} \tiny{\textcolor{gray}{$\pm$0.0035}} &
		0.9046 \tiny{\textcolor{gray}{$\pm$0.0046}} \\
	NBM$_{(500)}$ &
		\underline{0.9863} \tiny{\textcolor{gray}{$\pm$0.0011}} &
		0.9490 \tiny{\textcolor{gray}{$\pm$0.0043}} &
		\underline{0.8634} \tiny{\textcolor{gray}{$\pm$0.0017}} &
		0.8848 \tiny{\textcolor{gray}{$\pm$0.0006}} &
		0.7735 \tiny{\textcolor{gray}{$\pm$0.0068}} &
		0.9736 \tiny{\textcolor{gray}{$\pm$0.0022}} \\
    \bottomrule
  \end{tabular}
\end{table*}

\subsection{Baselines}
In addition to the NAM and NBM families, we evaluated six other models: EBM~\cite{ebm-implement}, NODE-GAM~\cite{NODE-GAM}, logistic regression (LR), decision tree (DT), MLP, and XGBoost~\cite{xgboost}. EBM and NODE-GAM are state-of-the-art GAMs and support to consider pairwise feature interactions. We denote EBM and NODE-GAM with feature interactions as EB$^2$M and NODE-GA$^2$M, respectively. It should be noted that EB$^2$M supports only regression and binary classification tasks. MLP and XGBboost are non-interpretable models but will be accurate. The hyperparameters of each model were tuned by grid search in the same manner as in the proposed model.

\subsection{Results}
\subsubsection{Comparison with Baselines}
Table \ref{tab:comp_baseline} lists the test accuracies of each model. The models in the first group belong to GAMs and only use univariate shape functions.
Among GAMs, NBM and NAM-FS$_{(500)}$ exhibit the best performance for two datasets. For high-dimensional datasets, guillermo ($D=4,296$) and Gisette ($D=5,000$), our NAM-FS$_{(500)}$ and NBM-FS$_{(500)}$ yield the highest accuracy among GAMs, respectively. In such high-dimensional datasets, NAM-FS$_{(500)}$ and NBM-FS$_{(500)}$ are also advantageous in terms of computational efficiency compared to vanilla NAM and NBM, as discussed in Section \ref{sec:complexity}. However, the performance of NAM-FS$_{(50)}$ and NBM-FS$_{(50)}$ is worse than that of other GAMs. This is because the number of features to be selected ($K_1=50$) is insufficient to obtain a better accuracy, while such a setting can speed up the training process and reduce the number of shape functions to be managed in high-dimensional datasets.

Focusing on the models belonging to GA$^2$Ms (second group in Table \ref{tab:comp_baseline}), either the proposed NA$^2$M-FS$_{(500, 500)}$ or NB$^2$M-FS$_{(500, 500)}$ exhibits the highest accuracy among GA$^2$Ms and outperforms state-of-the-art GA$^2$Ms. Even NA$^2$M-FS$_{(50, 50)}$ and NB$^2$M-FS$_{(50, 50)}$ that have only 50 shape functions for feature interactions are better than EB$^2$M and NODE-GA$^2$M on ISOLET and guillermo, which implies the potential of our feature selection. The user parameters $K_1$ and $K_2$ balance the trade-off between performance and computational cost. Setting these values as large as possible within the available computational resource will be a possible choice when pursuing performance.

Finally, checking the performance of other well-known models, logistic regression shows the best performance on Epsilon and Gisette, whereas it does not work well on F-MNIST and guillermo. Both MLP and XGBoost archive the best performance on two datasets out of six, which may be attributed to the complicated model architectures. Among the interpretable models (GAMs, GA$^2$Ms, logistic regression, and decision tree), the proposed NA$^2$M-FS$_{(500, 500)}$ and NB$^2$M-FS$_{(500, 500)}$ consistently achieve good performances, which demonstrates the effectiveness of our models.

\subsubsection{Evaluation of Feature Selection Mechanism} \label{sed:exp_presel}
We validated whether the feature-selection mechanism used in the proposed models works well. A naive approach for reducing the number of features to be considered is to pre-select the features before the training of NAM and NBM. We used mutual information (MI)-based feature selection implemented in the \texttt{scikit-learn} library~\cite{scikit-learn} and selected the top $K_1$ features according to the MI scores. We then trained NAM and NBM using only the selected $K_1$ features. We performed this two-stage method with $K_1=50, 500$, denoted by NAM$_{(K_1)}$ and NBM$_{(K_1)}$.

Table \ref{tab:prefs} shows the test accuracies of NAM and NBM with pre-selected features. In the case of $K_1=50$, we observe that our NAM-FS$_{(50)}$ and NBM-FS$_{(50)}$ are better than NAM and NBM with pre-selected features, except for guillermo. In addition, the performances of NAM and NBM with pre-selected features are greatly worse for the datasets other than guillermo. For $K_1=500$, NAM-FS$_{(500)}$ and NBM-FS$_{(500)}$ tend to be better than NAM$_{(500)}$ and NBM$_{(500)}$ for a large number of features ($D \leq 2,000$). This result demonstrates that our models can select appropriate features for NAM and NBM through end-to-end learning.

\section{Conclusion}
In this paper, we have proposed extensions of NAM and NBM by incorporating the feature selection mechanism in an end-to-end learning manner. We demonstrated the computational efficiency of the proposed models. Although the proposed models are simple, they allow us to incorporate pairwise feature interactions efficiently and scale for high-dimensional datasets. The experimental results demonstrate that our proposed models are better than or comparable to state-of-the-art GAMs. 

Although the proposed NAM-FS and NBM-FS enable us to reduce the NN shape functions in NAM and NBM and show promising performance among GAMs in high-dimensional datasets, the advantages of our models fade with fewer than 100 features. Another limitation is that we attempted to capture only pairwise interactions. It is possible to extend our NAM-FS and NBM-FS to consider more than two feature interactions, although this will compromise interpretability. An efficient implementation of NBM for sparse features was proposed in \cite{NBM}. Introducing such sparse feature support is a possible future work. Moreover, applying our models to real-world datasets, such as biological and medical datasets, is also an important direction.

\subsubsection*{Acknowledgements}
This work was partially supported by JSPS KAKENHI (JP20H04240, JP20H04254, JP22H03590, JP23H00491, JP23H03466), JST PRESTO (JPMJPR2133), NEDO (JPNP18002, JPNP20006), and a grant from the Kanagawa Prefectural Government of Japan.

\bibliographystyle{splncs04}
\bibliography{mybib}

\clearpage

\appendix

\section{Number of Parameters and MACs}
Table \ref{tab:params-macs} shows the number of parameters (\#param) and the total multiply-accumulate operations (MACs) in the forward calculation of NAM, NAM-FS, NBM, and NBM-FS. In the default setting used in the experiments, the number of parameters and MACs of one NN in NAM are $P_{\mathrm{NAM}}=M_{\mathrm{NAM}}=6,721$, and those in NBM are $P_{\mathrm{NBM}}=M_{\mathrm{NBM}}=64,044$. Comparing NAM and NAM-FS, with the number of features ($D$), \#param and MACs of NAM become $DP_{\mathrm{NAM}}$ and $DM_{\mathrm{NAM}}$, respectively, while they are $K_1(D+P_{\mathrm{NAM}})$ and $K_1(D+M_{\mathrm{NAM}})$ in NAM-FS. As $P_{\mathrm{NAM}} \gg K_1$ and $D>K_1$ in usual, \#param and MACs of NAM-FS are smaller than those of NAM as $D$ increases.

In NBM, \#param and MACs are $P_{\mathrm{NBM}} + DB$ and $D (M_{\mathrm{NBM}}+B)$, respectively. As $P_{\mathrm{NAM}} \gg B$ in usual, \#param in NBM does not become large compared to NAM as $D$ increases, while MACs increase as $D M_{\mathrm{NBM}}$ in a way similar to NAM. We note that MACs affect the calculation time of training and inference. This fact makes NBM not run efficiently on a large feature dimension and pairwise interactions. On the other hand, \#param and MACs of NBM-FS depend on $D$ as $DK_1$, which means that NBM-FS is advantageous for MACs in large $D$ because of $M_{\mathrm{NBM}} \gg K_1$.

For NA$^2$M and NB$^2$M, \#param and MACs can be roughly given by replacing $D$ with $D + D(D-1)/2$. For NA$^2$M-FS and NB$^2$M-FS, they can be given roughly by replacing $K_1$ with $K_1 + K_2$. This means that NA$^2$M-FS is more efficient than NA$^2$M with respect to both \#param and MACs, and NB$^2$M-FS is more efficient than NB$^2$M with respect to MACs.

Table \ref{tab:params-macs-na2m} shows the actual \#param and MACs in a specific setting with $D=500$. The architectures of the NNs are the same as those used in the experiment. We observed that \#param of NA$^2$M and MACs of NA$^2$M and NB$^2$M are greater than those of our models by orders of magnitude. We note that the larger NN is used in NBM in the default setting~\cite{NBM}, which is why the MACs of NB$^2$M are greater than that of NA$^2$M. Therefore, it is practically difficult to run NA$^2$M and NB$^2$M training in high-dimensional features using a usual computing resource.

\begin{table}[tb]
  \centering
  \caption{Comparison of the number of parameters (\#param) and the total multiply–accumulate operations (MACs) in forward calculation of each model with a single output. The number of parameters of a NN used in NAM and NBM is represented by $P_{\mathrm{NAM}}$ and $P_{\mathrm{NBM}}$, respectively. The MACs of a NN used in NAM and NBM are represented by $M_{\mathrm{NAM}}$ and $M_{\mathrm{NBM}}$, respectively. Other parameters are as follows: $D$: the number of features, $B$: the number of bases in NBM, $K_1$: the number of one-input shape functions in NAM-FS and NBM-FS. We ignored the scalar bias term $b$ for simplicity.}
  \label{tab:params-macs}
  \begin{tabular}{lll}
    \toprule
    & \#param & MACs \\
    \midrule
    NAM     & $DP_{\mathrm{NAM}}$ & $DM_{\mathrm{NAM}}$ \\
    NAM-FS  & $K_1(D+P_{\mathrm{NAM}})$ & $K_1(D+M_{\mathrm{NAM}})$ \\
    NBM     & $P_{\mathrm{NBM}} + DB$ & $D(M_{\mathrm{NBM}}+B)$ \\
    NBM-FS  & $P_{\mathrm{NBM}} + K_1(D+B)$  & $K_1(D+M_{\mathrm{NBM}}+B)$ \\
    \bottomrule
  \end{tabular}
\end{table}

\begin{table}[tb]
  \centering
  \caption{Comparison of \#param and MACs in $D=500$. The numbers of one- and two-input shape functions in NA$^2$M-FS and NB$^2$M-FS are set to $K_1=K_2=500$, respectively.}
  \label{tab:params-macs-na2m}
  \vspace{0.5\baselineskip}
  \begin{tabular}{lrr}
    \toprule
    & \#param & MACs \\
    \midrule
    NA$^2$M & 849.8M & 849.8M \\
    NA$^2$M-FS & 7.5M & 7.5M \\
    NB$^2$M & 25M & 9,719M \\
    NB$^2$M-FS & 0.98M & 65M \\
    \bottomrule
  \end{tabular}
\end{table}

\section{Datasets}
We use the following six classification datasets in the experiments. For F-MNIST and Epsilon, we use the test data provided in the datasets. We randomly split the data into 80\% and 20\% for training and testing, respectively, for other datasets.
We obtained the HAR, ISOLET, guillermo, Gisette datasets from the OpenML website.\footnote{\url{https://www.openml.org/}}

\subsubsection{HAR~\cite{HAR}} The task of this dataset is the human activity recognition (HAR) built from 30 subjects performing activities of daily living while carrying a waist-mounted smartphone with embedded inertial sensors. The task is to classify each sensor signal feature vector into six activities (walking, walking upstairs, walking downstairs, sitting, standing, and laying).

\subsubsection{ISOLET~\cite{ISOLET}} This dataset consists of pre-processed speech data from 150 subjects. Each subject spoke the name of each letter of the English alphabet. This dataset is widely used to evaluate feature selection methods.

\subsubsection{F-MNIST~\cite{FMNIST}} Fashion MNIST (F-MNIST) is a different version of MNIST, which consists of $28 \times 28$ grayscale images of clothing items from 10 classes. We treat the $784$ $(=28 \times 28)$ pixels as the feature vector.

\subsubsection{Epsilon~\cite{Epsilon}} The task of this dataset is binary classification, which was used in the PASCAL Large Scale Learning Challenge 2008.

\subsubsection{guillermo~\cite{Guillermo}} The guillermo dataset of binary classification is used in the AutoML challenge.

\subsubsection{Gisette~\cite{gisette}} The Gisette is used for the feature selection competition held in NIPS 2003. The task is based on a handwritten digit recognition problem to separate the highly confusable digits `4' and `9'. This dataset has the highest number of features in this work.

\section{Hyperparameter Setting and Tuning}
We tune the hyperparameters of each model using a grid search. We randomly select 10\% from the training dataset as a validation dataset and train the models with candidate hyperparameters using the remaining 90\% training dataset. Then, we select the best-performing hyperparameters for validation datasets with respect to accuracy. After deciding on the best hyperparameters, we train each model ten times with different random seeds and report the average test accuracy.

NAM and NBM families, NODE-GAM, and MLP were run using a single NVIDIA A100 GPU (40GB or 80GB memory), and EBM, LR, DT, and XGboost were run using only CPUs.

\subsubsection{NAM, NA$^2$M, NAM-FS, and NA$^2$M-FS}
We tune the dropout rate from $\{ 0.0, 0.2, 0.5 \}$, the feature dropout rate from $\{ 0.0, 0.05 \}$, the weight decay coefficient from $\{0.0, 10^{-4}\}$, and the output penalty coefficient from $\{0.0, 0.01, 0.1 \}$. Model architectures and training procedures are described in the main content. We use an early stopping strategy in which we stop the training if the validation accuracy does not improve for 11,000 iterations and decay the learning rate to 1/5 if the validation accuracy does not improve for 5,000 iterations. We terminate the training if the training iteration reaches 100K and monitor the validation performance every 500 iterations. We use the PyTorch~\cite{pytorch} library (version 1.13.1) to implement these models.

\subsubsection{NBM, NB$^2$M, NBM-FS, and NB$^2$M-FS}
We tune the dropout rate from $\{ 0.0, 0.2, 0.5 \}$, the basis dropout rate from $\{ 0.0, 0.5 \}$, the weight decay coefficient from $\{0.0, 10^{-4}\}$, and the output penalty coefficient from $\{0.0, 0.01, 0.1 \}$. Model architectures are described in the main content. The training procedure is the same as in NAMs. We use the PyTorch~\cite{pytorch} library (version 1.13.1) to implement these models.

\subsubsection{EBM}
We use the \texttt{interpret} library\footnote{\url{https://github.com/interpretml/interpret}} as the implementation of the EBM. We tune the outer bags from $\{ 4, 8, 16 \}$ and the inner bags from $\{ 0, 4, 8 \}$. Other hyperparameters are set as default values.

\subsubsection{NODE-GAM}
We use the official implementation of NODE-GAM.\footnote{\url{https://github.com/zzzace2000/nodegam}}. The early stopping strategy is the same as in NAMs and NBMs. We tune the number of trees per layer from $\{ 200, 500 \}$, the number of layers of trees from $\{2, 3, 4\}$, the depth of the tree from $\{ 2, 4, 6\}$, and the architecture from $\{ \mathrm{GAM}, \mathrm{GAMAtt} \}$. Other hyperparameters are set as default values.

\subsubsection{Logistic Regression (LR)}
We use \texttt{scikit-learn} library~\cite{scikit-learn} (version 1.2.1). We tune the penalty term from $\{ \mathrm{L1}, \mathrm{L2} \}$, and the inverse of the regularization strength from $\{ 0.01, 0.1, 1, 10, 100, 1000 \}$.

\subsubsection{Decision Tree (DT)}
We use \texttt{scikit-learn} library~\cite{scikit-learn} (version 1.2.1). We tune the split criterion from $\{ \mathrm{gini}, \mathrm{entropy} \}$, the maximum depth of the tree from $\{ 4, 8, 16, 32 \}$, the minimum number of samples required to split an internal node from $\{ 2, 8, 16 \}$, and the minimum number of samples required to be at a leaf node from $\{ 1, 8, 16 \}$.

\subsubsection{Multi-Layer Perceptron (MLP)}
We tune the architecture from (i) 5 hidden layers with 128, 128, 64, 64, 64 units; (ii) 3 hidden layers with 1024, 512, 512 units; (iii) 2 hidden layers with 2048, 1024 units. These candidate architectures are referred from \cite{NBM}. We apply batch normalization followed by ReLU activation in each layer. We also tune the dropout rate from $\{ 0.0, 0.2, 0.5 \}$ and the weight decay coefficient from $\{0.0, 10^{-4}\}$. The training procedure is the same as in NAMs and NBMs.

\subsubsection{XGBoost}
We use \texttt{xgboost} library.\footnote{\url{https://xgboost.readthedocs.io/en/stable/}} We tune the number of boosting rounds from $\{ 100, 200, 500, 1000 \}$, the maximum tree depth from $\{ 2, 4, 8, 16 \}$, the subsample ratio of columns from $\{ 0.5, 0.8, 1.0 \}$, and the subsample ratio of the training instance from $\{ 0.5, 0.8, 1.0 \}$. We use the early stopping strategy using the validation dataset and set the early stopping rounds to 50. Other hyperparameters are set as default values.

\section{Illustration of Model Architectures}
\begin{figure}[!tb]
    \centering
    \includegraphics[keepaspectratio, width=0.65\columnwidth]{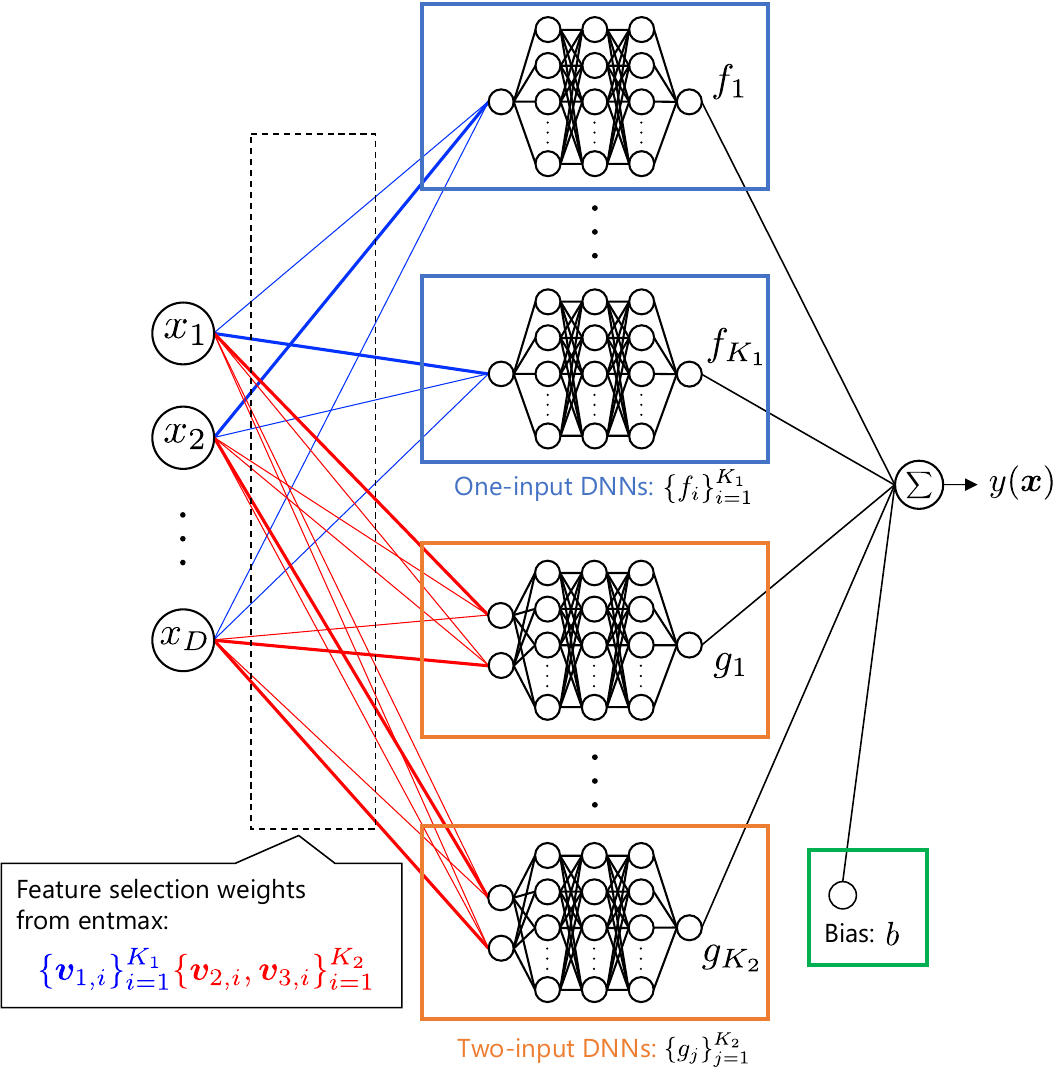}
    \caption{Model architecture of NAM-FS for single output.}
    \label{fig:nam_img}
\end{figure}

\begin{figure}[!tb]
    \centering
    \includegraphics[keepaspectratio, width=0.8\columnwidth]{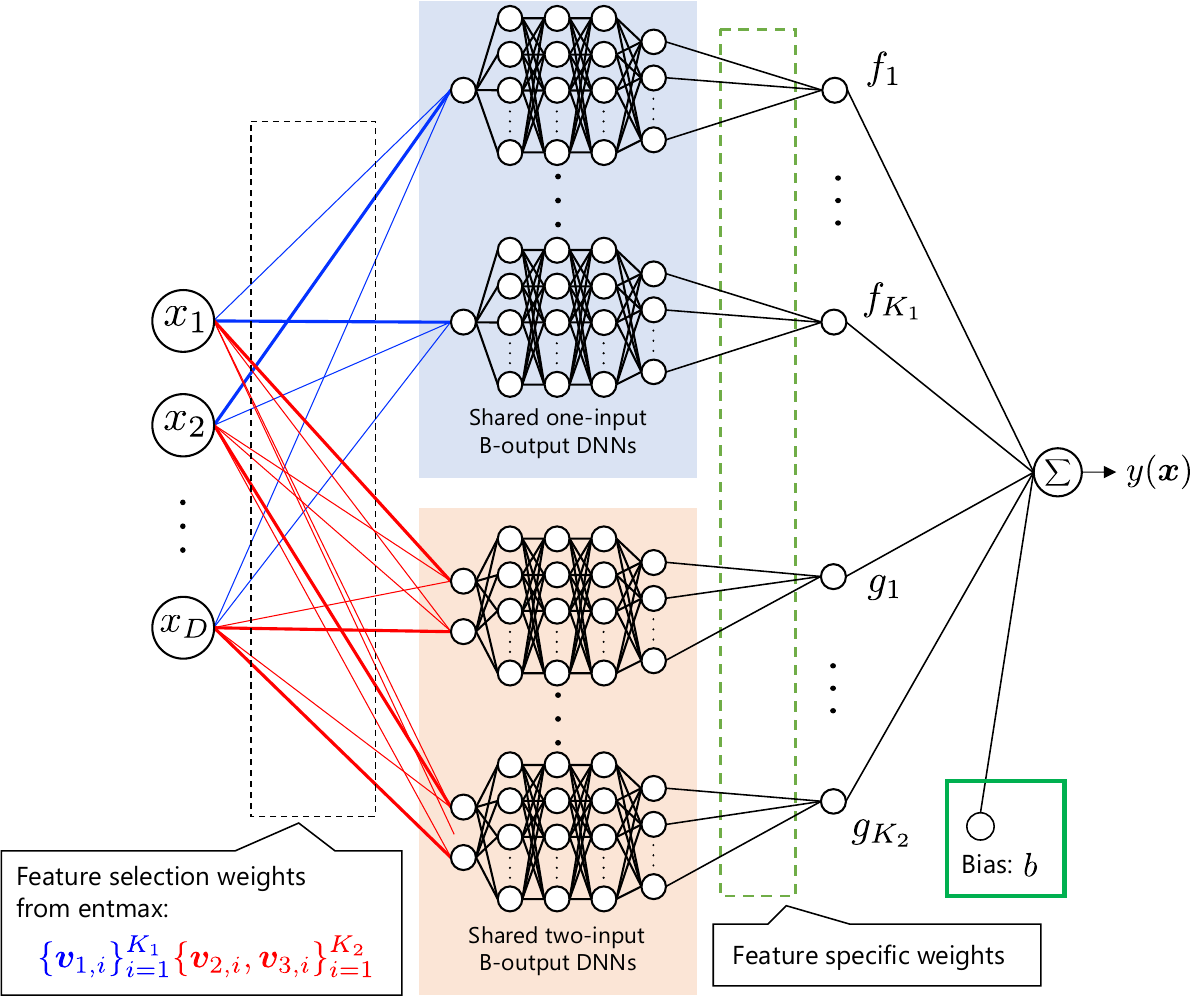}
    \caption{Model architecture of NBM-FS for single output.}
    \label{fig:nbm_img}
\end{figure}

Figures \ref{fig:nam_img} and \ref{fig:nbm_img} illustrate the model architectures of our NAM-FS and NBM-FS, respectively.

\end{document}